# Public Sentiment Toward Solar Energy: Opinion Mining of Twitter Using a Transformer-Based Language Model

Serena Y. Kim[a,b,*], Koushik Ganesan[c], Princess Dickens[d] and Soumya Panda[e]

[a]*Computer Science, University of Colorado Boulder, 1111 Engineering Dr, Boulder, CO 80309 USA*
[b]*School of Public Affairs, University of Colorado Denver, 1380 Lawrence St., Suite 500, Denver, CO 80204 USA*
[c]*Physics, University of Colorado Boulder, 2000 Colorado Ave, Boulder, CO 80309 USA*
[d]*Linguistics, University of Colorado Boulder, Hellems 290, Boulder, CO 80309 USA*
[e]*Business Analytics, University of Colorado Boulder, 995 Regent Dr, Boulder, CO 80309 USA*

ABSTRACT

Public acceptance and support for renewable energy are important determinants of renewable energy policies and market conditions. This paper examines public sentiment toward solar energy in the United States using data from Twitter, a micro-blogging platform in which people post messages, known as tweets. We filtered tweets specific to solar energy and performed a classification task using Robustly optimized Bidirectional Encoder Representations from Transformers (RoBERTa). Analyzing 71,262 tweets during the period of late January to early July 2020, we find public sentiment varies significantly across states. Within the study period, the Northeastern U.S. region shows more positive sentiment toward solar energy than did the Southern U.S. region. Solar radiation does not correlate to variation in solar sentiment across states. We also find that public sentiment toward solar correlates to renewable energy policy and market conditions, specifically, Renewable Portfolio Standards (RPS) targets, customer-friendly net metering policies, and a mature solar market.

## 1. Introduction

For the first time in U.S. history, solar energy recently became more cost efficient than coal energy in the long term [36]. Solar energy contributes to greenhouse gas (GHG) emission reduction, offers new business opportunities to landowners and energy providers, and allows utility customers to lower their utility bills. Despite these advantages, only about 2% of U.S. electricity generation currently emanates from solar [64]. The low proportion of energy generated from solar may be attributed to multiple factors, including skepticism surrounding financial benefits, system reliability, and conflicts of interest with respect to utility revenue preservation [73].

Although public sentiment and preferences regarding renewable energy have been studied in the last decade [43, 59, 20], few studies have documented public perception of solar energy specifically. Public opinion on solar energy is worth investigating separately because different renewable sources such as geothermal, hydroelectricity, wind, and bioenergy have unique advantages and requirements depending on the existing climate conditions and state-specific energy policies.

Furthermore, most of the existing literature relies on surveys and interviews to understand public perception of renewable energy. Although surveys and interviews can provide targeted and relevant data, they can be susceptible to selection and response biases [9]. For example, people who are more supportive of renewable energy may be more likely to respond to surveys and interviews, which may yield bias. Social media is an increasingly popular means to express opinions and preferences and can thus provide valuable information for understanding public opinion on solar energy.

In this paper, our first aim is to understand public opinion using data from Twitter, a micro-blogging platform in which people can post and interact with messages known as "tweets". With over 50 million active users in the United States alone [2], Twitter provides an ideal platform for opinion mining as extensive amounts of data are collected across a wide range of demographics and geographical locations. To achieve this aim, we utilized Robustly optimized

*Corresponding author

serena.kim@colorado.edu (S.Y. Kim); koushik.ganesan@colorado.edu (K. Ganesan); princess.dickens@colorado.edu (P. Dickens); soumya.panda@colorado.edu (S. Panda)

ORCID(s): 0000-0003-3839-3674 (S.Y. Kim)

Bidirectional Encoder Representations from Transformers (RoBERTa) [38] based on recent development in the fields of Natural Language Processing (NLP) and Machine Learning (ML). RoBERTa possesses an extensive pre-training phase which can later be fine-tuned for a domain-specific task, yielding highly accurate results. Our second aim in this paper is to examine whether the energy policies and market conditions explain public opinion on solar energy. In particular, we focus on state-level characteristics, including renewable energy generation capacity, renewable energy portfolio standards (RPS), net metering, renewable energy incentives, and solar market maturity.

The results suggest that public opinion on solar energy varies widely across states and is more likely to be positive in states with aggressive RPS targets, customer-friendly net metering rules, and a more mature solar market. This study has policy implications in addressing spatial disparities in public support for solar energy and future opportunities for solar energy deployment. In particular, states which implemented specific renewable energy incentives are more supportive of solar energy, which may have implications for solar deployment. We also discuss the benefits of using social media as a source of data on public opinion and applying RoBERTa for sentiment classification in gauging public sentiment toward solar energy.

## 2. Background

Public acceptance and awareness of renewable energy are critical to the development of renewable energy industries and technology [52]. On one hand, lack of institutional, political, and community support is often identified as a key barrier to renewable energy development [73]. Landscape modification and visual intrusion of facilities are highlighted as main reasons for residents' opposition to renewable energy [6, 71]. On the other hand, the contribution of renewable energy to the local and regional economy positively influences public support for renewable energy [47, 7]. Not surprisingly, individuals' opinion and preferences regarding renewable energy are highly associated with their belief and perception of global warming, climate change, and environmental risk [16, 43]. Views on renewable energy are also determined by personal characteristics such as education, party identification, and age [20, 56].

Government policies and public opinion on renewable energy may relate to each other in a bidirectional sense. Public acceptance and support for renewable energy may affect renewable energy policy adoption, which in turn encourages renewable energy deployment. Renewable energy programs and financial incentives can also mitigate uncertainties in energy transition, garnering public support and acceptance for renewable energy. For example, RPS policy design and framing strongly influence broad public support for renewable energy technologies [59]. Additionally, net metering, or an electricity billing policy allowing renewable energy system owners (e.g., residential, commercial, and industrial buildings) to use the renewable energy generated, has been shown to facilitate solar energy deployment [12, 13]. A well-designed, transparent net metering policy can mitigate market and revenue uncertainties and help gain support from key stakeholders, including utilities, solar businesses, and customers [5], despite compensation rate limitations negatively impacting solar energy preferences after reaching a certain tipping point [12].

Public beliefs about the effectiveness of the policy-making and planning process is an important predictor of public sentiment toward renewable energy. Public perceptions of fairness in decision making about the siting of renewable energy facilities can mitigate public resistance to renewable energy deployments [71]. Having trust in the policy makers responsible for renewable energy development has a direct effect on public opinion [25]. Greater openness to information sharing about alternative energy options can enhance public support for renewable energy [78]. However, lack of a common understanding about the planning process contributes to public reluctance of renewable energy adoption [8]. Public education and outreach around administrative and technological aspects of solar energy can help reverse negative perceptions of renewable energy development [53, 25]. Leveraging social networks and facilitating participation in the planning and development process can also enhance public trust in solar energy development [44].

Electricity market characteristics and conditions also likely influence public opinion on renewable energy. Public preferences depend on the price of conventional electricity because renewable energy is considered an alternative to conventional electricity generation. While the availability of low cost fossil fuel-generated electricity can make it difficult to justify renewable energy development [10], increased energy prices can enhance public acceptance of renewable energy [61]. In addition, high upfront costs and lack of adequate financing options are major barriers to public support for renewable energy [44, 25]. Solar businesses contribute to local economic development, by lowering solar installation costs and creating new jobs, which in turn leads to public support for renewable energy development.

Existing literature also documents temporal and spatial variations in preferences and opinions regarding renewable energy. Public opinion about renewable energy changes over time [66, 1]. For example, Hamilton et al. [20] finds that public acceptance of renewable energy shows clear upward trends from 2011 to 2018. Historical events (e.g., energy shortages and electricity price increases) can affect public awareness about renewable energy technology [1]. Public acceptance of renewable energy also varies geographically, by communities [11], states [37, 20], and countries [58]. Spatial and geographical characteristics, such as region-specific culture [4], solar radiation [57], and energy autonomy of the region [58], influence spatial variations in public opinion.

Most empirical studies have used surveys and interviews to measure public opinion, sentiment, awareness, and perception of renewable energy. The literature finds broad public support for renewable energy across the United States [59], Finland [33], Mexico [19], Spain [54], South Korea [29], Portugal [60], Greece [24], and worldwide [52, 72]. Surveys and interviews have advantages in gauging individual-level demographic information, such as gender, education, income [34], distance to renewable energy facilities, and previous experience with renewable energy technologies [58], which is one of the key determinants of individuals' preferences regarding renewable energy.

However, surveys and interviews are limited in gauging temporal dynamics and geographical variations in public opinions. Researchers are beginning to use social media, especially Twitter, to examine public sentiment toward renewable energy [1, 23, 46, 37]. Jain and Jain [23] compares five different machine learning techniques for sentiment analysis and finds that the Support Vector Machine (SVM) achieves higher accuracy than K-Nearest Neighbor, Naive Bayes, AdaBoost, and Bagging algorithms for sentiment classification on renewable energy related tweets. Using both traditional and social media for opinion mining, Nuortimo and Härkönen [46] find that public opinion on solar and wind has been the most positive compared to other energy sources, including coal, nuclear, and biomass. Using Twitter data from 2014 to 2016, Abdar et al. [1] finds that Alaskans' energy preferences have become more supportive of renewable energy over time.

## 3. Materials and methods

Our data are from two main sources: Twitter [63] and the U.S. federal and state government agencies. We use Twitter data for sentiment analysis, or opinion mining, of solar energy. Twitter has been a valuable source for opinion mining, but the manual classification of a large amount of tweets is difficult and time-consuming. Thus, we use NLP and ML methods to detect public opinion on solar energy automatically. We discuss our data collection and pre-processing processes in the following section. Section 3.2 outlines recent developments and approaches in opinion mining. Section 3.3 explains our sentiment classification model built upon RoBERTa. Finally, section 3.4 summarizes energy policy and market data used in this study.

### 3.1. Twitter data collection and pre-processing

Twitter Application Program Interface (API) was used to collect tweets, which are posts created by individuals on Twitter, specific to solar energy. Tweepy, a python library for accessing the Twitter API is used to stream live tweets in real-time [63]. We used ten keywords to stream live tweets, including 'solar energy', 'solar panel', 'solar PV', 'solar photovoltaic', 'solar battery', 'solar thermal', 'solar power', 'solar-powered', 'solar generation', and 'solar subsidies.' In total, 406,811 tweets specific to solar energy were collected between January and early July of 2020.

We removed URLs, "RT (ReTweets)", and images from original tweets using the library preprocessing [41], which aids processing text strings. We identified a list of words that make a tweet irrelevant to public sentiment toward solar energy. These words include 'Pokemon', 'Superman', 'galaxy', 'eclipse', 'solar plexus', 'solar-powered human', and 'I will become your sun.' We excluded 16,245 tweets that included these words. We also excluded the tweets that included the 10 key words (e.g., 'solar energy', 'solar panel', 'solar PV', 'solar photovoltaic') only in the user identification (i.e., screen name and description) but not in the text, quoted text, or extended text. We excluded 64,253 tweets through this process.

Not all tweets include geographic information that is essential to this study. Only about 40% of the tweets have geographic information either where they are based or where users tweeted. Among this 40%, about half of the tweets are from countries other than the United States. For the purpose of this study, we only need the tweets tweeted by users with geolocations associated with the United States and/or the tweets by the users who identify themselves based in the U.S. Thus we extracted tweets with geographic information based on the self-reported locations in the user profiles as well as the latitude/longitude coordinates using Carmen, a library for geolocating tweets [15]. Our final dataset includes 71,262 unlabelled tweets as a result of this extraction process.

We randomly selected 5,122 of the 71,262 tweets and manually classified them into one of the two groups: "positive toward solar" or "negative toward solar." Tweets that include job announcements and product advertisement in the solar industry are classified into the positive class as these tweets support solar energy industry. Retweets are included since retweeting is a way to express individuals' support or opposition to solar energy. We did not include a neutral class because there were not enough tweets (8% of collected tweets) which we would consider neutral. In this case, including a neutral class could have resulted in lower performance or prediction accuracy [68, 48]. The tweets classified as neutral were deleted from the manually annotated dataset(5,122 tweets). Table 1 illustrates examples of tweets belonging to the two classes, positive and negative solar sentiment.

**Table 1**
Sample Tweets for Two Classes

| Tweets | Sentiment |
|---|---|
| Solar energy has never been easier and more affordable to install.<br><br>Oil is the the way of the past.<br>Solar and other forms of clean energy is where you have to invest.<br><br>Add solar and get paid twice – Equity and Energy! | Positive |
| I've seen gopher tortoises and red-shouldered hawks rendered bereft of habitat because of solar panel farms.<br><br>Solar and wind power are totally dependent on weather and can't be trusted.<br><br>Solar is expensive to maintain and return is not what everyone is shouting about. | Negative |

## 3.2. Related opinion mining approaches

ML approaches have been widely applied to opinion mining, also known as sentiment analysis. Opinion mining is the computational treatment of opinions, sentiments, and subjectivity of text [28]. Sentiment classification techniques based on NLP and ML have been widely applied for opinion mining in diverse areas, including public health [79], movie reviews [27], airline service [67], political news [26], and the COVID-19 pandemic [17, 40]. Sentiment classification is a sub-discipline of text classification, which is concerned with classifying a text to a class for analyzing opinion or sentiment in texts. Although human emotions and intents are highly complex in real life, the current state-of-the-art sentiment analysis has achieved higher performance with a simpler classification task, such as classifying texts into two (e.g., positive and negative) or three categories (e.g., positive, negative, and neutral).

Recent developments in NLP have produced effective ways for automatic sentiment analysis using ML algorithms [50]. Lexicon-based approaches using supervised and unsupervised ML algorithms, including Naïve Bayes and Support Vector Machines (SVM), have generated high accuracy in sentiment classification. They have gained wide popularity in opinion mining [21]. More recently, deep learning approaches including Gated Recurrent Units (GRUs), long short-term memory (LSTM), and Convolutional Neural Networks (CNN) have started using non-static word embedding and have shown even higher performance results than the previous ML approaches in multiple sentiment classification tasks [39, 49].

Although neural network models have established higher accuracy with their automated feature learning, the application of such models is often limited by their heavy reliance on manually annotated data for training language models [50]. Within neural network models, considerable progress has been achieved by the models using the Transformer architecture, which is based on an self-attention mechanism [65]. While such advantages have given rise to the development of new models, BERT (Bidirectional Encoder Representations from Transformers) which uses contextual sentences and word-embeddings to address the limitations of RNNs and LSTMs, has notably improved the performance of sentiment analysis [14]. BERT is also the first large-scale bidirectional unsupervised way of training the language model by using two main strategies, namely as "masked language model" (MLM) and "next sentence prediction". A number of BERT-based models have been developed since 2018 including DistilBERT [55], AlBERT [35], XLNet [74], and RoBERTa[38]. One of the key advantages of the BERT-based models is that users do not need a large corpus of texts to train their models. Users only need to fine-tune a BERT-based model using area/task-specific supervised training data (e.g., manually annotated tweets) because BERT is already pre-trained on large corpus from Wikipedia and books, while utilizing more rich contextual information. The sentence-level vectors are more convenient for downstream NLP tasks, including sentiment classification, detecting sentence similarity, and learning word vector embedding.

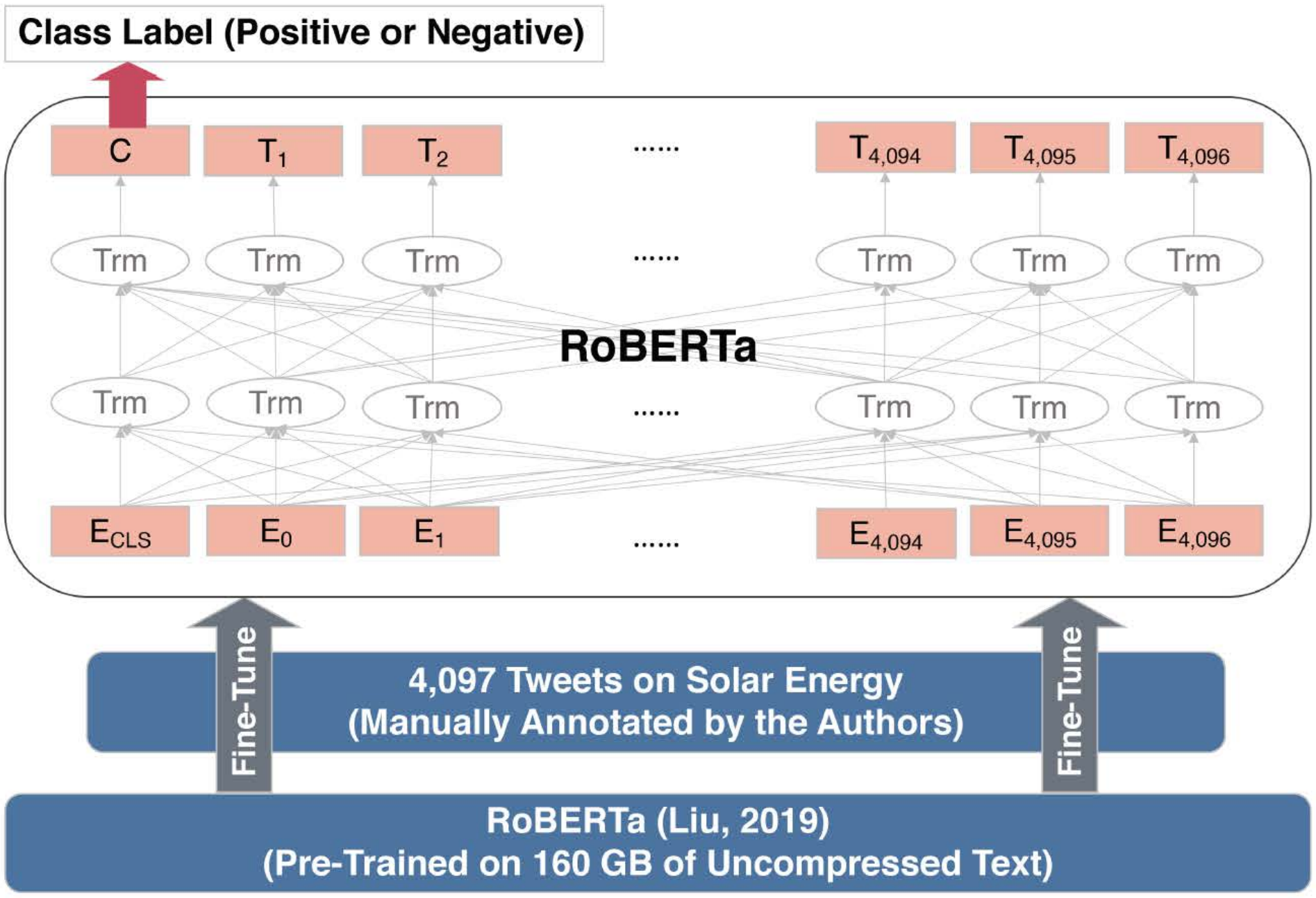

**Figure 1:** Graphic representation of our model based on RoBERTa. The input embeddings are denoted as E, and the final hidden vector is denoted as T. Every input example starts with CLS, a special token. C is the final hidden vector of the special CLS token. Trm is Transformer. Adapted from Devlin et al. [14].

## 3.3. Our approach: RoBERTa-based sentiment classification

Our sentiment classification model is based on RoBERTa, a Robustly Optimized BERT [38]. Like any other BERT-based models, RoBERTa is powerful as it pretrains the language model by a bidirectional representation of words, meaning that the model is not restricted to reading texts either right-to-left or left-to-right. This deep bidirectional pre-training structure provides the model with more information about context. RoBERTa also achieves higher accuracy than all other previous BERT-based models including BERT, BERT-Large, and XLNet. This higher accuracy is achieved by four main modifications, including pre-training the language model with 8-times larger batches over 10-times more data; training on 5-times longer sequences; using Byte-Pair Encoding (BPE) vocabulary instead of the character-level vocabulary; removing the next sentence prediction (NSP); dynamically changing the masking pattern applied to the training data instead of static masking [38]. Thus, we chose RoBERTa as our base model, as it is considered the state-of-the-art sentiment classification approach as of January 2020.

In total, we manually annotated 5,122 tweets without duplicates, and the annotated tweets were used to construct our train (80%), development (10%) and test (10%) sets. As shown in Figure1, we fine-tuned RoBERTa using 4,097 annotated tweets. During the annotation process, we detected sarcastic expressions (e.g., "Solar power? Yeah, right. That is a top concern for millions of Americans during COVID".) and classified the tweets including such expressions into negative class.

We used the HuggingFace transformer library [70], which includes the standard RoBERTa-based architecture with 12 hidden states and 12 attention heads, but with some modifications to the hyper-parameters, the parameters in which values are used to control the learning process in machine learning. We also used the AdamW optimizer [32] to minimize the cross-entropy loss. To optimize the parameters, we conducted a number of experiments testing multiple combinations of the hyper-parameters on the development set (10% of the annotated tweets). In our final model, we fine-tuned RoBERTa with a learning rate of 6e-6, an $\epsilon$ of 1e-8, and dropout of 0.1. We set the maximum sequence length to 128 tokens, used a batch size of 16 and train for 4 epochs on a Tesla P100 GPU. Using this final set of hyper-parameters, our model achieves 90.7% accuracy with an F1 score of 0.927 on the test set (another 10% of annotated tweets).

## 3.4. Renewable energy policy and market

The second aim of this study is to examine associations between public sentiment on solar energy and renewable energy policy and market characteristics. The characteristics examined in this study include renewable energy generation, RPS, net metering, the number of existing renewable energy incentives and policies, electricity price, and solar energy market maturity.

### *3.4.1. Renewable generation*

A state's existing renewable energy generation capacity may be positively associated with public acceptance of solar energy. Renewable generation is measured as the percentage of renewable energy (excluding hydroelectric energy) generated in a state in Megawatt-hours as of 2019. Electricity generation data are from the U.S. Energy Information Administration (EIA) [64]. Distributed energy generation (e.g., rooftop solar) is not included in the dataset due to a lack of nationwide data on distributed energy generation capacity.

### *3.4.2. RPS*

RPS policy is a state-level mechanism that requires utilities to generate or purchase a certain percentage of energy from renewable sources. Although RPS policies share common characteristics, the design features vary widely across states. Existing studies have developed and used different RPS measures, including a binary measure of whether the state has RPS policies [75], a RPS percentage target [22], an interaction between the percentage target and the target year[31], marginal RPS targets [30], and a RPS target combined with "free trade" of Renewable Energy Certificates (RECs) [77]. In order to take into account such variations, we construct a measure for RPS as follows:

$$RPS_i = \frac{TargetPercent_i - 2019Generation_i}{TargetYear_i - 2019} \tag{1}$$

States that already achieved a RPS target or do not have RPS policies as of 2019 received a score of 0. RPS data as of 2019's Quarter 4 are from state government agencies. Iowa and Texas do not have a target year, but both states already achieved RPS target and received a score of 0.

### *3.4.3. Net metering*

Net metering data are from the North Carolina Clean Energy Technology Center [3]. Some states have enacted net metering policies and other distributed generation (DG) compensation rules to allow electricity customers generating electricity (e.g., rooftop solar) to use that generated energy at any time. Solar system owners send the excessive energy back to the electricity grid and are compensated at a specific rate. DG compensation policies provide substantial incentives for solar energy generation as commercial and residential building owners can use solar energy at night, generated during the sunny or cloudy days. Net metering policies can vary significantly in terms of design and features. Thus, we construct an additive measure capturing five key features:

a. $Mechanism_i$: The existence of statewide net metering mechanisms (4 = statewide net metering; 3 = statewide alternative compensation mechanism; 2 = some customers (e.g., residential buildings) receive net metering benefits; 1 = only selective utilities (e.g., IOUs) provide net metering; 0 = no net metering or alternative DG compensation)
b. $Cap_i$: Net metering capacity limitations, which regulate the size of systems which can receive net metering benefits in states (1 = unlimited system size; 0 = otherwise)
c. $Subscriber_i$: Net metering subscriber size limitation (1 = unlimited; 0 = otherwise)
d. $Compensation_i$: Compensation rate for energy generation (1 = compensate for customer rates; 0 = otherwise)
e. $Rollover_i$: Rollover of remaining energy is allowed (2 = allowed without any limitations; 1 = partially allowed or allowed only until the end of billing year; 0 = not allowed)

$$NEM_i = Mechanism_i + Cap_i + Subscriber_i + Compensation_i + Rollover_i \tag{2}$$

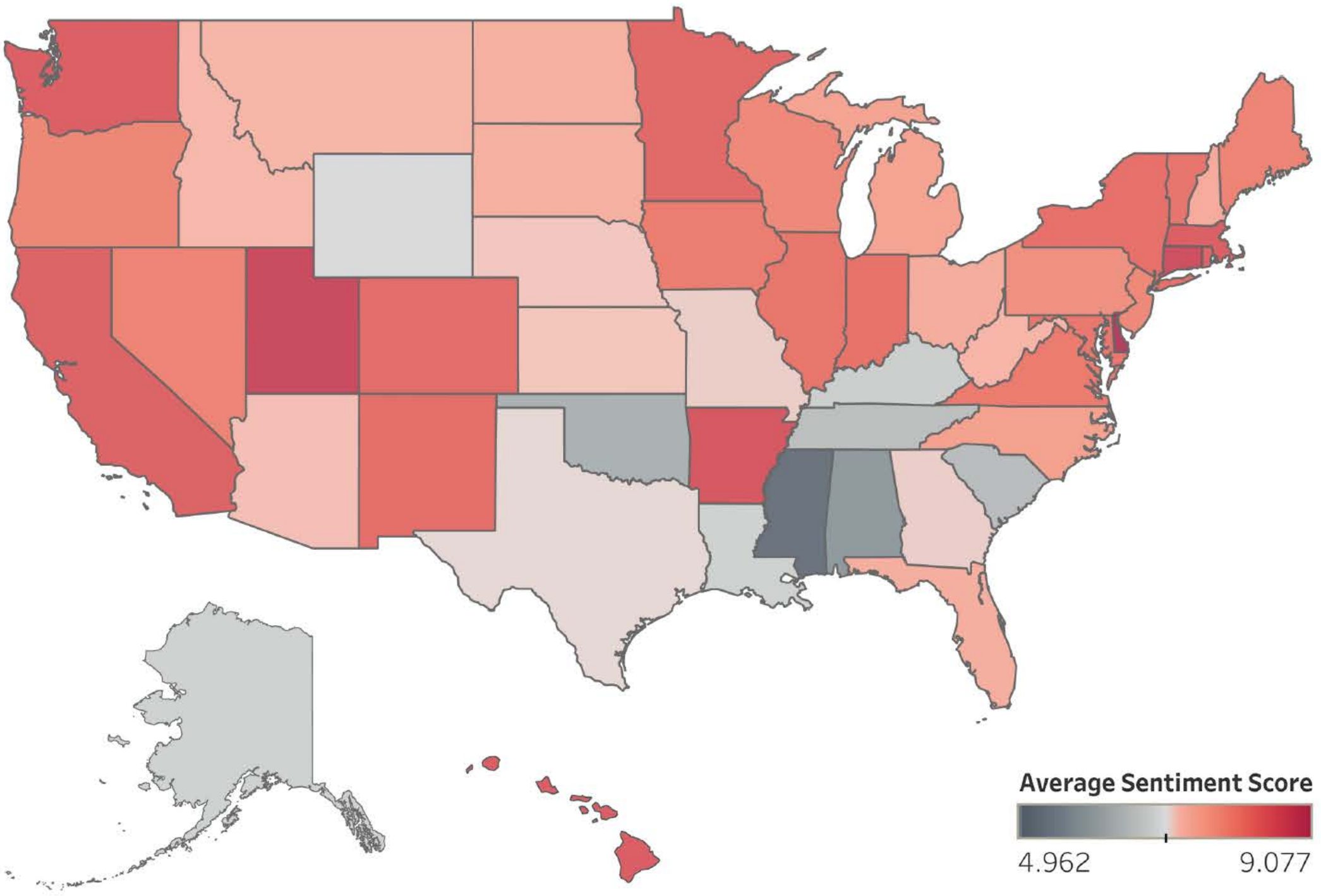

**Figure 2:** Spatial patterns in sentiment toward solar energy across the United States

### *3.4.4. Renewable incentives*

States which enacted more renewable energy policies and incentives may have greater public support for renewable energy. Thus, we included the number of renewable energy policies and incentives by states as of October 2019. The data were collected from the database of State Incentives for Renewable & Efficiency [42].

### *3.4.5. Solar market maturity*

Public sentiment on solar energy may be more positive in states with a more mature solar market as this creates jobs and supports local economic development. We measure solar market maturity by the number of solar industry jobs per 1 million people in the state population in 2019. The data were obtained from the National Solar Job Census [62].

### *3.4.6. Electricity price*

Higher electricity prices may be associated with higher public sentiment toward solar energy because distributed renewable energy generation helps reduce electricity bills in places with high electricity rates [61, 10]. Thus, we include in our analysis the average price of electricity to ultimate residential customers by the state as of August 2019. The electricity data are from the U.S. Energy Information Administration (EIA) [64].

### *3.4.7. Solar radiation*

The level of solar radiation positively predicts solar photovoltaic system installations [57] and solar energy generation [22]. As the solar energy system achieves greater efficiency and performance in regions with high solar radiation, we expect that annual average solar radiation is positively associated with sentiment toward solar energy. Solar radiation is measured in KWh/$m^2$/Day and is aggregated at the state level. The solar radiation data are from the National Renewable Energy Laboratory's National Solar Radiation Database [45].

**Table 2**
Descriptive statistics

| Variables | Obs. | Mean | SD | Min | Max | (1) | (2) | (3) | (4) | (5) | (6) | (7) |
|---|---|---|---|---|---|---|---|---|---|---|---|---|
| (1) Average sentiment score | 51 | 7.546 | .805 | 4.96 | 9.08 | 1 | | | | | | |
| (2) Renewable generation | 51 | 17.078 | 17.079 | 1 | 67 | .13 | 1 | | | | | |
| (3) RPS | 51 | 1.568 | 1.825 | 0 | 9 | .40 | -.06 | 1 | | | | |
| (4) Net metering | 51 | 6 | 2.088 | 0 | 9 | .43 | .07 | .12 | 1 | | | |
| (5) Renewable incentives | 51 | 68.961 | 41.386 | 13 | 217 | .28 | .07 | .16 | .09 | 1 | | |
| (6) Solar market maturity | 51 | .725 | .695 | 0 | 2 | .44 | .09 | .24 | .18 | .23 | 1 | |
| (7) Electricity price | 51 | 13.863 | 4.441 | 10 | 33 | .26 | -.04 | .26 | .17 | .02 | .34 | 1 |
| (8) Solar radiation | 51 | 4.333 | .589 | 3 | 6 | -.22 | .08 | -.14 | -.13 | .03 | .18 | -.11 |

# 4. Results

## 4.1. Public opinion on solar energy by state

Figure 2 presents the sentiment score findings across the United States. The District of Columbia (9.077), Utah (8.7 71), and Connecticut (8.692) are most supportive of solar energy, while Mississippi (4.962), Alabama (5.745), and Oklahoma (6.224) are least supportive. There is a statistically significant difference (one-way ANOVA, $F(3,47) = 3.30$, $p = .028$) across four U.S. Census regions (Northeast, Midwest, South, and West), but the Bartlett's test with Bonferroni adjustment indicates that only the difference between the Northeast and the South is statistically significant ($\overline{Northeast} - \overline{South} = .975, p = .042$).

Figure 3 displays average sentiment scores of 50 states and the District of Columbia (n = 51) between January 12 and July 7, 2020. The average sentiment score can vary from 0 (all tweets are negative toward solar energy) to 10 (all tweets are positive toward solar energy). The state average sentiment score ranges from 4.962 (Mississippi) to 9.077 (District of Columbia). The 95% confidence intervals depend on the number of tweets collected from each state. The number of tweets from each state varies from 64 (North Dako ta) to 11,788 (California), with an overall mean of 1397.294 and standard deviation of 1967.161, resulting in larger confidence intervals for North Dakota, South Dakota, and Wyomi ng. However, the number of tweets per 1 million people in the state population has a lower standard deviation of 307.08 7 for a mean of 242.996, indicating that the average number of tweets adjusted for state population is more evenly distributed.

## 4.2. Sentiment toward solar energy and renewable energy policy and market characteristics

Table 2 presents the descriptive statistics for all variables included in the statistical analysis. The average sentiment score of each state is calculated by averaging sentiment scores of all tweets from each state. Thus, the mean of the average sentiment scores (7.546) in table 2 is slightly different from the national average (7.835), as the means for states are based on non-weighted state averages. The highest correlation between the policy and market variables is .34, between solar market maturity and electricity price.

Table 3 presents the results from the regression of sentiment score on the renewable energy policy and market variables. In models 1–7, we examine bivariate associations between sentiment and each renewable energy policy/market variable. The full model (model 8) includes all predictor variables. Model 1 demonstrates that there is no significant relationship between public sentiment and current renewable energy generation. In model 2, public sentiment on solar energy, however, is more positive in states which enacted higher annual RPS targets ($\beta = .178$, $p < .01$). We also find that states with more customer-friendly net metering policies (e.g., statewide net metering mechanisms, no capacity limitations, year-to-year rollover) have a more positive sentiment toward solar energy ($\beta = .167$, $p < .01$) in model 3. This finding is consistent in the full model ($\beta = .115$, $p < .01$). Model 3 suggests that net metering explains 18.8% of the variation in public sentiment on solar energy.

The number of renewable energy incentives in states is positively correlated with public sentiment in model 4 ($\beta = .005$, $p < .05$), but this relationship is not statistically significant in the full model. In model 5, we find that solar energy is perceived more positively in states with a more mature solar market ($\beta =.508$, $p < .01$). This finding also holds in the full model ($\beta = .373$, $p < .01$). The results from model 6 suggest average sentiment score is higher in states with higher electricity price ($\beta = .046$, $p < .05$), but this association is not statistically significant in the full model. Lastly, solar radiation is not associated with public sentiment on solar energy. The full model has no multicollinearity issues (mean VIF = 1.15) and explains 46.9% of the total variance in the sentiment score.

| State | No. of Tweets | Score |
|---|---|---|
| District of Columbia | 2,274 | 9.077 |
| Utah | 773 | 8.771 |
| Conneticut | 879 | 8.692 |
| Arkansas | 207 | 8.502 |
| Hawaii | 402 | 8.483 |
| Washington | 2,116 | 8.426 |
| Massachusetts | 1,733 | 8.413 |
| California | 11,788 | 8.380 |
| Rhode Island | 127 | 8.346 |
| Minnesota | 1,414 | 8.260 |
| Colorado | 2,375 | 8.248 |
| New Mexico | 523 | 8.222 |
| New York | 5,650 | 8.191 |
| Indiana | 1,320 | 8.167 |
| Illinois | 2,171 | 8.125 |
| Vermont | 325 | 8.123 |
| Maryland | 1,021 | 8.100 |
| Virginia | 1,641 | 8.062 |
| Iowa | 469 | 8.038 |
| Nevada | 1,180 | 7.941 |
| Maine | 359 | 7.939 |
| Oregon | 1,466 | 7.906 |
| Delaware | 181 | 7.901 |
| New Jersey | 1,328 | 7.877 |
| Wisconsin | 906 | 7.859 |
| Pennsylvania | 1,554 | 7.780 |
| North Carolina | 1,794 | 7.547 |
| Michigan | 1,552 | 7.519 |
| New Hampshire | 301 | 7.409 |
| Ohio | 1,936 | 7.392 |
| Florida | 5,072 | 7.376 |
| South Dakota | 87 | 7.356 |
| North Dakota | 64 | 7.344 |
| West Virginia | 216 | 7.269 |
| Montana | 255 | 7.255 |
| Idaho | 333 | 7.237 |
| Arizona | 1,851 | 7.175 |
| Kansas | 631 | 7.147 |
| Nebraska | 255 | 7.137 |
| Missouri | 722 | 7.105 |
| Georgia | 1,473 | 7.101 |
| Texas | 5,989 | 7.058 |
| Wyoming | 111 | 6.937 |
| Louisiana | 532 | 6.805 |
| Kentucky | 548 | 6.770 |
| Alaska | 213 | 6.761 |
| Tennessee | 1,025 | 6.488 |
| South Carolina | 589 | 6.435 |
| Oklahoma | 527 | 6.224 |
| Alabama | 738 | 5.745 |
| Mississippi | 266 | 4.962 |

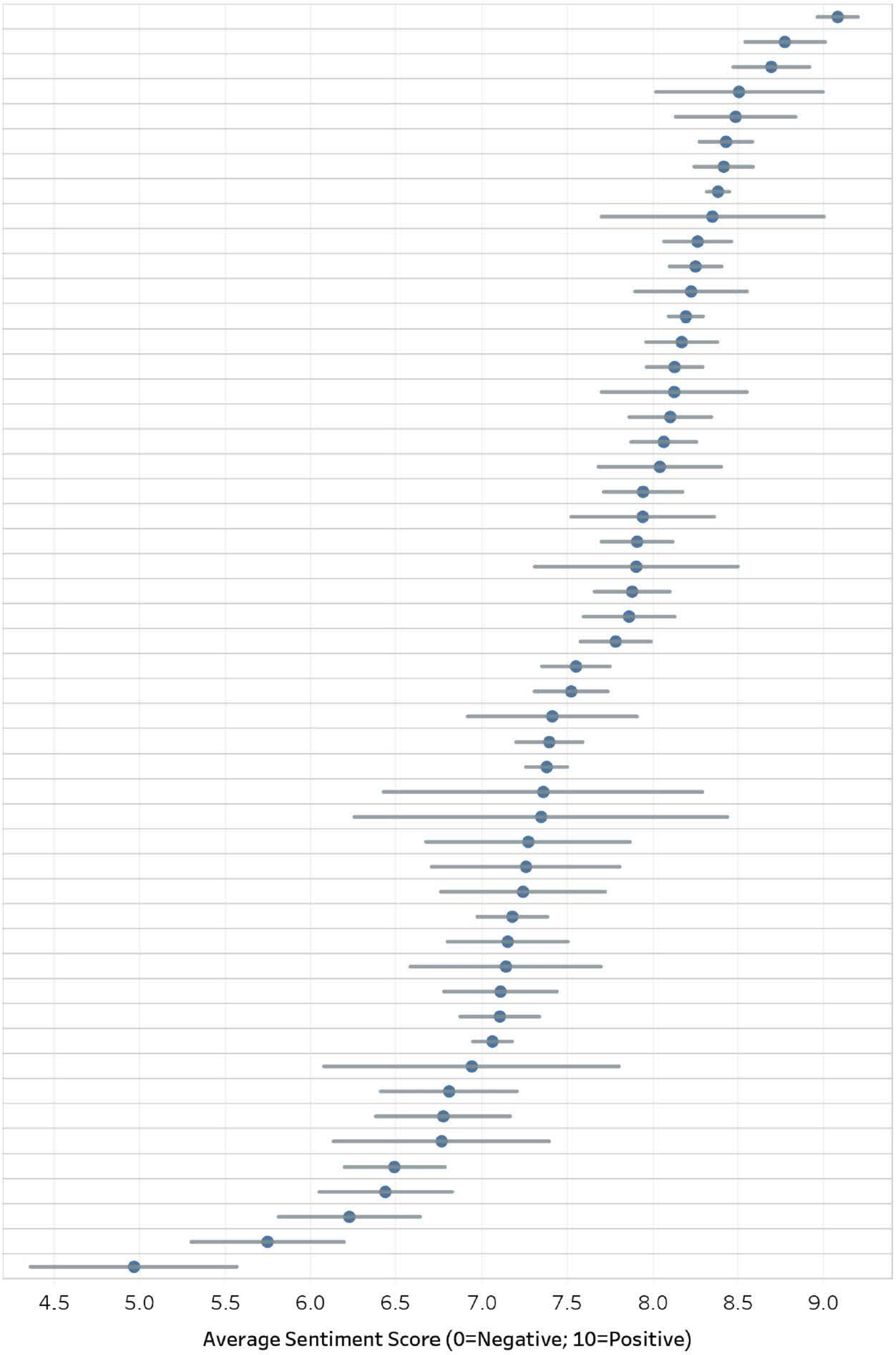

**Figure 3:** Average sentiment score in each state (n=71,262; national average=7.835; error bars indicate 95% confidence intervals)

**Table 3**
Public sentiment toward solar energy and renewable energy policy and market characteristics

| Variables | (1) | (2) | (3) | (4) | (5) | (6) | (7) | (8) |
|---|---|---|---|---|---|---|---|---|
| Renewable generation | .006 | | | | | | | .005 |
| | (.006) | | | | | | | (.005) |
| RPS | | .178*** | | | | | | .106*** |
| | | (.058) | | | | | | (.029) |
| Net metering | | | .167*** | | | | | .115*** |
| | | | (.048) | | | | | (.037) |
| Renewable incentives | | | | .005** | | | | .003 |
| | | | | (.002) | | | | (.002) |
| Solar market maturity | | | | | .508*** | | | .373*** |
| | | | | | (.129) | | | (.119) |
| Electricity price | | | | | | .046** | | .002 |
| | | | | | | (.021) | | (.019) |
| Solar radiation | | | | | | | -.301 | -.293 |
| | | | | | | | (.221) | (.156) |
| Number of states | 51 | 51 | 51 | 51 | 51 | 51 | 51 | 51 |
| $R^2$ | .016 | .163 | .188 | .079 | .193 | .066 | .048 | .469 |

***$p < 0.01$, **$p < 0.05$, *$p < 0.1$. Robust standard errors are in parentheses.

## 5. Discussion

This paper aimed to understand how sentiment toward solar energy varies across states, and how policy and market factors explain suh sentiment. The results indicate overall sentiment toward solar energy is positive but substantial variation exists across states. As of early 2020, the Northeast region is most supportive of solar energy, while the South region is the lowest. Solar radiation, which determines the efficiency and performance of the solar energy generation system, does not explain geographical variation in public sentiment toward solar, although this may be a result of such radiation varying more widely in states with the largest land area. But we find that energy policies and solar market maturity explain the variation in sentiment. This finding suggests that state energy policy programs could be effective measures to build public support for renewable energy.

We also detect temporal variation in sentiment for the 6-month period of Twitter data collection. As shown in Figure 4, events can potentially shift public opinion on solar energy. The sharp decline in sentiment score on March 23–25 may be motivated by the congress' discussion on the inclusion of tax incentives for renewable energy in the Coronavirus Aid, Relief, and Economic Security (CARES) Act of 2020. To ensure the robustness of our results, we employed an alternative model excluding the tweets from March 23–25. As shown in Table A1, the main findings from the alternative model were consistent with the original model, except for the estimated effect of renewable incentives. In the alternative model, renewable incentives are statistically significant in explaining solar sentiment in both the bivariate and full models. The alternative model also explains more of the total variation ($R^2 = .499$).

Our analysis finds the proportion of energy generated from renewable sources does not relate to public sentiment toward solar energy, but the RPS targets appear to matter. That is, public opinion on solar energy aligns more closely with state renewable energy policy goals but not current renewable energy generation. Although our analysis does not examine causal relationships, these findings suggest public opinion and policy may affect each other: public support for solar energy may impact RPS policy adoption and design, but at the same time, a progressive RPS policy may help build public support for solar energy. The design and framing of RPS policies may shape public support for renewable energy [59].

Our analysis also finds that net metering policy is also a strong predictor of public opinion on solar energy. As net metering offers direct financial benefits to residential and commercial solar system owners, we suspect that state net-metering policies influence public support for solar energy. Considering the impact of net metering rules and retail rates on consumer-side solar system deployment [12], specific design elements (e.g., compensation rate, capacity limitations, subscriber size limitations), as opposed to simply having net metering rules, may play an important role in fostering positive public perceptions of solar energy. For example, indefinite rollover of excess renewable generation credits and compensation at retail electricity prices provides more explicit incentives, which in turn may promote positive

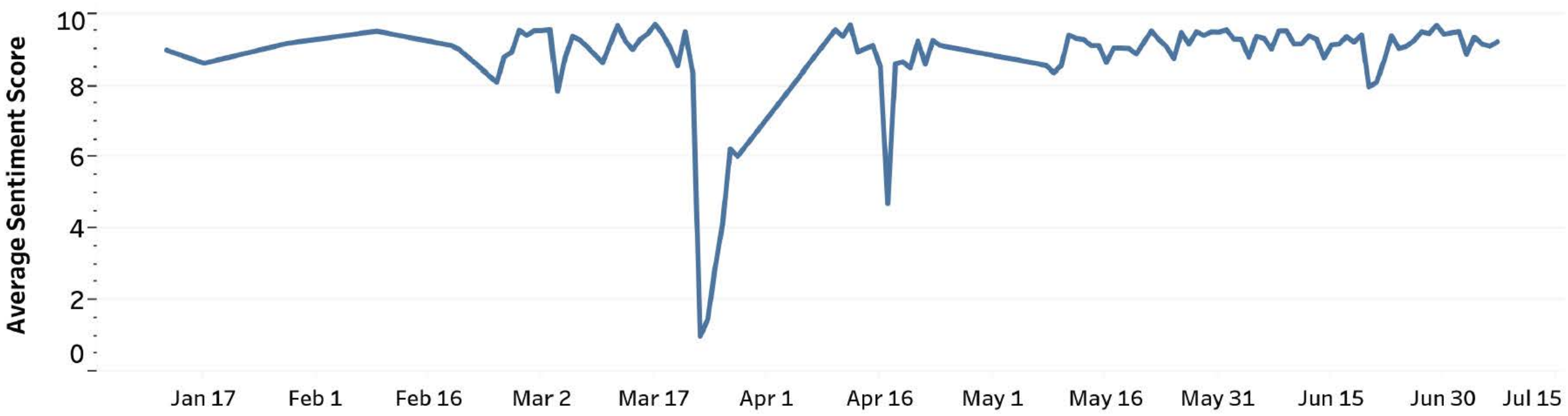

**Figure 4:** Trend of public sentiment toward solar energy from January 12, 2020 to July 07, 2020

perception of solar energy.

Solar market maturity, measured by the number of solar jobs per 1 million residents in the state population, is one of the most important predictors of public sentiment toward solar energy. This finding suggests public support of solar energy could positively influence solar energy market growth, and vice versa. Public acceptance of solar energy may be one of the factors that attracts solar businesses. At the same time, economies of scale in the solar market could enable solar firms to lower system installation costs and grid fees, generating "trickle down" effects that benefit solar energy customers and increasing public support for solar energy.

This study has a few key limitations. First, the Twitter data we collected may not be fully representative of all U.S. residents due to demographic disparities between Twitter users and non-users [2]. Twitter users tend to be younger and more politically liberal than the general public [69], and at the same time, younger and more liberal populations tend to favor renewable energy development [20]. Therefore, our estimation of the average sentiment score (7.835) is likely to be inflated compared to the true sentiment score. Considering sentiment classification can be highly dependent on the platform from which the training data are extracted, future research could address this limitation by incorporating data from multiple social media platforms.

Second, our analysis captures public sentiment toward solar energy between late January and early July 2020, amid a global pandemic, which may limit the generalizability of the findings. Although the number of tweets used to estimate the average sentiment score is over 70,000, and the findings are robust to inclusion/exclusion of outlier values from March 23–25, the sentiment score could be made less sensitive if the data were from a longer period of time. In future research, we aim to collect data from multiple years and explore the sequential order of public sentiment and renewable energy policies to answer the question: Does public perception of solar energy precede renewable energy policies, or vice versa?

Finally, although our language model achieves high accuracy (90.7%) with binary classification (positive or negative), it is not sufficient to capture fine-grained human emotions, such as happiness, joy, excitement, anger, sadness, frustration, fear, and sarcasm. As proposed by Abdar et al. [1], public preferences and attitudes on renewable energy can be gauged with multiple dimensions, including valence (positive versus negative) and arousal (high versus low level of activation). Future research should aim to capture multidimensional emotions, for example, by utilizing SentiBert to capture compositional sentiment semantics [76], modeling irony and sarcasm detection [51], and/or explicitly modeling label semantics in guiding an encoder network [18].

## 6. Conclusions

This study aimed to understand public sentiment toward solar energy in the United States and examine the relationship between sentiment and renewable energy policy and market characteristics. We collected 71,262 tweets from late January to early July 2020 and analyzed the data using RoBERTa, a state-of-the-art language model. We find that public sentiment on solar energy varies widely across states. RPS and net metering policies and solar market maturity are important predictors of the public sentiment toward solar energy.

The main contributions of this study are as follows:

i. We examine public opinion and sentiment specific to solar energy that has not been separately investigated as

extensively by existing literature.

ii. This study is one of few empirical studies using social media, ML, and NLP approaches to understand public opinion and sentiment regarding renewable energy.

iii. Applying RoBERTa, which is a state-of-the-art language model as of early 2020, we propose a new method to examine public opinion and sentiment toward solar energy. Our sentiment classification model achieves high (90.7%) accuracy by using manually annotated tweets, which are area-specific and of higher quality, to fine-tune RoBERTa.

iv. This study provides a comprehensive picture of the geographical variation in public sentiment regarding solar energy across states. We demonstrate that this variation is explained by state policy and market characteristics while refuting the theory that it can be explained by local solar radiation amounts.

v. Our analysis offers important policy implications as we provide empirical evidence of the positive relationship between public sentiment toward solar energy and renewable energy policy and market characteristics. States that wish to gain public support for solar energy may need to consider implementing consumer-friendly net metering policies and support state solar market growth.

## CRediT authorship contribution statement

**Serena Y. Kim:** Conceptualization of the study, Language modeling, Data collection, Statistical and spatial analysis, Writing-original draft. **Koushik Ganesan:** Neural network modeling, Language modeling, Writing-review and ed-iting. **Princess Dickens:** Language modeling, Data curation, Writing-review and editing. **Soumya Panda:** Language modeling, Data curation, Writing-review.

## Acknowledgement

The authors thank Heeyoung Kwon, Katharina Kann, and William Swann for helpful suggestions.

## Appendix

**Table A1**

Public sentiment toward solar energy and renewable energy policy and market characteristics (Excluding outlier values from March 23–25, 2020)

| Variables | (1) | (2) | (3) | (4) | (5) | (6) | (7) | (8) |
|---|---|---|---|---|---|---|---|---|
| Renewable generation | .004 | | | | | | | .002 |
| | (.005) | | | | | | | (.005) |
| RPS | | .146*** | | | | | | .082*** |
| | | (.049) | | | | | | (.029) |
| Net metering | | | .120*** | | | | | .081** |
| | | | (.045) | | | | | (.034) |
| Renewable incentives | | | | .005*** | | | | .003** |
| | | | | (.002) | | | | (.002) |
| Solar market maturity | | | | | .457*** | | | .333** |
| | | | | | (.100) | | | (.124) |
| Electricity price | | | | | | .045*** | | .011 |
| | | | | | | (.016) | | (.015) |
| Solar radiation | | | | | | | -.165 | -.164 |
| | | | | | | | (.178) | (.125) |
| Number of states | 51 | 51 | 51 | 51 | 51 | 51 | 51 | 51 |
| $R^2$ | .013 | .177 | .158 | .106 | .249 | .099 | .048 | .499 |

***$p < 0.01$, **$p < 0.05$, *$p < 0.1$. Robust standard errors are in parentheses.